\pdfoutput=1

\documentclass[11pt]{article}

\usepackage[]{acl}

\usepackage{times}
\usepackage{latexsym}

\usepackage[T1]{fontenc}

\usepackage[utf8]{inputenc}

\usepackage{microtype}

\usepackage{inconsolata}
\usepackage{multirow}
\usepackage{multicol}
\usepackage{graphicx}
\usepackage{array}
\usepackage{graphicx}
\usepackage{microtype}
\usepackage{amsfonts}
\usepackage{mathrsfs}
\usepackage{subfigure}
\usepackage{amssymb}
\usepackage{bbding}
\usepackage{subfigure} 
\usepackage{CJKutf8}
\usepackage{colortbl}
\usepackage{color}
\usepackage{booktabs}
\usepackage{makecell}
\usepackage{calligra}
\usepackage{algorithm} 
\usepackage{subcaption}
\usepackage{subfigure}
\usepackage{amsmath}

\title{Joint Information Extraction Across Classical and Modern Chinese with Tea-MoELORA}

\author{
  Xuemei Tang\textsuperscript{1},\quad
  Chengxi Yan\textsuperscript{2}\thanks{Corresponding authors.},\quad
 Jinghang Gu\textsuperscript{1}\footnote[1],\quad
  and Chu-Ren Huang\textsuperscript{1}\\
  \textsuperscript{1}The Hong Kong Polytechnic University, Hong Kong, China \\
  \textsuperscript{2}Renmin University of China, Beijing, China \\
  \texttt{xuemeitang00@gmail.com} \\
}

\begin{document}
\begin{CJK}{UTF8}{gbsn}  

\maketitle
\begin{abstract}
Chinese information extraction (IE) involves multiple tasks across diverse temporal domains, including Classical and Modern documents. Fine-tuning a single model on heterogeneous tasks and across different eras may lead to interference and reduced performance. Therefore, in this paper, we propose Tea-MoELORA, a parameter-efficient multi-task framework that combines LoRA with a Mixture-of-Experts (MoE) design. Multiple low-rank LoRA experts specialize in different IE tasks and eras, while a task-era-aware router mechanism dynamically allocates expert contributions. Experiments show that Tea-MoELORA outperforms both single-task and joint LoRA baselines, demonstrating its ability to leverage task and temporal knowledge effectively.
\end{abstract}

\section{Introduction}

 Typical information extraction (IE) tasks consist of Named Entity Recognition (NER), Relation Extraction (RE) and Event Extraction (EE). With recent advances in large language models (LLMs), IE performance has greatly improved. To further leverage LLMs, a number of studies have explored unified fine-tuning approaches that integrate multiple IE tasks into a single framework~\cite{Gui_Zhang_Ye_Zhang_2023,Wang_Zhou_Zu_Xia_Chen_Zhang_Zheng_Ye_Zhang_Gui_et_2023}. These efforts are motivated by the observation that IE tasks often share overlapping features and representations.

However, despite the potential for shared learning, naively mixing heterogeneous data—especially across different task types and time periods—can lead to task conflicts and feature interference, ultimately degrading model performance~\cite{Liu_Wu_Zhao_Zhu_Xu_Tian_Zheng_2024, Feng_Hao_Zhang_Han_Wang_2024, Dou_Zhou_Liu_Gao_Shen_Xiong_Zhou_Wang_Xi_Fan_etal_2024}. This issue is particularly pronounced in Chinese IE, where corpora often exhibit strong temporal characteristics (e.g., classical Chinese historical texts vs. modern Chinese social media).
These observations highlight the need to revisit the fine-tuning paradigm for Chinese IE in light of heterogeneous task formats and temporal variability. Specifically, we identify two central challenges that must be addressed: 
\textbf{(i) Task Diversity:} Chinese IE involves various tasks (e.g., NER, RE, EE, temporal extraction) that differ in input length and output format. Training separate models is resource-intensive, and while multi-task learning (MTL)~\cite{Gui_Zhang_Ye_Zhang_2023} offers a unified approach, most methods overlook the impact of temporal variation.
\textbf{(ii) Temporal Domain Shift:} Modern and historical Chinese IE tasks are typically treated separately~\cite{bao-etal-2024-employing, Tang_Wang_Wang_2026}, yet their linguistic continuity suggests shared features. Ignoring this connection can lead to underutilization of cross-era knowledge~\cite{Tang_Su_2022, Wei_Li_Zhu_Xu_Li_Wu_2025}. These challenges indicate that a straightforward multi-task fine-tuning approach may fail to fully exploit the shared structure across tasks and eras. To address both task diversity and temporal domain shifts, recent methods have explored combining parameter-efficient adapters with modular architectures that can dynamically allocate expertise based on input characteristics.

Recently, Mixture of LoRA Experts has been proposed to address multi-task learning challenges~\cite{Wu_Huang_Wei_2024, Liu_Wu_Zhao_Zhu_Xu_Tian_Zheng_2024}. This approach combines the LoRA~\cite{Hu_Shen_Wallis_Allen-Zhu_Li_Wang_Wang_Chen_2021} fine-tuning method with the MoE~\cite{6797059} framework. Multiple LoRA modules are designed as experts to capture both shared and task-specific knowledge, while a router mechanism dynamically adjusts each expert’s contribution.
Inspired by this, we adopt a unified training process to conduct parameter-efficient fine-tuning across multiple tasks. Only a small set of low-rank matrices are trained while freezing the backbone LLM, which helps mitigate both task diversity and temporal variation. 

Based on this idea, we propose Tea-MoELORA (Task-Era-Aware Mixture-of-Experts LoRA), a unified framework for multi-task fine-tuning that effectively incorporates both task- and era-specific knowledge through multiple LoRA experts and a dynamic routing mechanism. Rather than relying on a single LoRA module for all tasks and eras—which may lead to negative transfer—we employ multiple experts implemented as low-rank adapters. Each expert is specialized for different information extraction subtasks and temporal domains. A task-era-aware router dynamically assigns weights to these experts according to the input’s task identity and era label (e.g., classical vs. modern Chinese). This design enables instance-level, tailored parameter updates, achieving both task adaptability and era sensitivity within a single unified training process.

In the end, our contributions can be summarized as follows.

\begin{itemize}
    \item We introduce Tea-MoELORA, a novel joint information extraction framework that integrates the strengths of MoE and LoRA to achieve enhanced performance.
    \item Our framework integrates not only different IE tasks, such as relation extraction and event extraction, but also corpora from different historical eras, including Classical Chinese and Modern Chinese.
    \item Experimental results demonstrate the effectiveness of our framework, showing clear advantages over LoRA trained either individually or jointly without MoE.
\end{itemize}



\section{Related Work}

\subsection{Chinese Information Extraction}
Generative language models are employed to handle various IE tasks by converting their outputs into a universal format~\cite{Fei_Wu_Li_Li_Li_Qin_Zhang_Zhang_Chua_2022, Lu_Liu_Dai_Xiao_Lin_Han_Sun_Wu_2022, Wang_Zhou_Zu_Xia_Chen_Zhang_Zheng_Ye_Zhang_Gui_et_2023,Gui_Zhang_Ye_Zhang_2023}. This unified generation paradigm allows different information extraction tasks, such as NER, RE, and EE, to be formulated as conditional sequence generation problems. This flexibility simplifies model architecture and promotes transferability across tasks. For example, 
InstructUIE~\cite{Wang_Zhou_Zu_Xia_Chen_Zhang_Zheng_Ye_Zhang_Gui_et_2023} is an end-to-end framework for general information extraction, which leverages multi-task instruction tuning of LLMs to perform various IE tasks within a unified interface.
~\citet{Guo_Li_Jin_Liu_Zeng_Liu_Li_Yang_Bai_Guo_et_2023} proposed Code4UIE, a universal retrieval-augmented code generation framework for IE tasks. It defines task-specific schemas using Python classes and transforms information extraction into code generation with in-context learning guidance. 
~\citet{Wei_Cui_Cheng_Wang_Zhang_Huang_Xie_Xu_Chen_Zhang_et_2024} proposed a two-stage information extraction framework named ChatIE, which performs zero-shot IE through multi-turn question answering.

\subsection{LoRA with MoE}

Recent advances have explored integrating LoRA with MoE to improve multi-task or multi-domain adaptation.

Building upon multiple pre-trained LoRA modules, ~\citet{Xu_Lai_Huang_2024} introduced MeteoRA (Multiple-tasks embedded LoRA), which reuses existing LoRA adapters and performs token-level MoE routing through a learnable gating mechanism, which dynamically selects experts for each token based on the output of each LoRA adapter.

~\citet{Lin_Fu_Liu_Li_Sun_2024} applied the MoE architecture at the adapter layer and uses a gated unit driven by task description prompts to estimate relevance between source and target tasks, generating sparse adapter combinations tailored for transfer.
Similar works, such as LoraHub~\cite{Huang_Liu_Lin_Pang_Du_Lin_2024}, LoraRetriever~\cite{Zhao_Gan_Wang_Zhou_Yang_Kuang_Wu_2024}.

To enable simultaneous training of multiple LoRA modules, each serving as an expert,
~\citet{Feng_Hao_Zhang_Han_Wang_2024} proposed Mixture-of-LoRAs (MoA), where domain-specific LoRA modules serve as experts. A sequence-level routing strategy is designed based on domain metadata, allowing each transformer layer to dynamically select the appropriate expert during training.
Similarly,
~\citet{Wu_Huang_Wei_2024} treated each trained LoRA as an expert and integrates a layer-wise gating mechanism to learn adaptive combination weights, allowing efficient and flexible reuse of LoRA modules across tasks.
~\citet{Zhou_Li_Tian_Ni_Liu_Ye_Chai_2024} proposed SilverSight, a financial multi-task learning framework that clusters training data based on semantic similarity. Expert models are then aligned with the most semantically related tasks, forming a task-aware multi-expert system.
To address catastrophic forgetting in domain-specific adaptation, ~\citet{Dou_Zhou_Liu_Gao_Shen_Xiong_Zhou_Wang_Xi_Fan_etal_2024} proposed LoRAMoE, a framework that freezes the base model and incorporates multiple LoRA experts via a routing network. Some experts are designated to preserve world knowledge, mitigating catastrophic forgetting during domain-specific adaptation.

Rather than using full LoRA modules as experts, ~\citet{Liu_Wu_Zhao_Zhu_Xu_Tian_Zheng_2024} introduced MoELORA, where each expert is a part of low-rank matrices. A task-motivated gating mechanism is applied uniformly across all MoELORA layers, enabling efficient parameter usage while maintaining task specialization.
Observing that the parameters of matrix A trained on different tasks are highly similar, ~\citet{Tian_Shi_Guo_Li_Xu} proposed an asymmetric LoRA framework that enables sharing matrix A across tasks, thereby reducing parameter redundancy.
~\citet{lin-etal-2025-teamlora} introduced TeamLoRA, a novel framework that organizes and shares knowledge by interpreting matrix A as a general, domain-agnostic knowledge network and matrix B as a task-specific knowledge network. ~\citet{zhou-etal-2025-cola} presented CoLA, which assigns different numbers of experts to matrices A and B, respectively. A many-to-many collaborative strategy is designed to enable more fine-grained and flexible incremental updates.

Despite these advances, challenges persist in settings involving heterogeneous and temporally distant data—for instance, bridging Classical and Modern Chinese. These domains differ significantly in lexicon, grammar, and discourse, posing unique obstacles to expert routing and representation learning even though they share a historical lineage.

\section{Methodology}
\begin{figure*}[t]
    \centering
    \includegraphics[width=15cm, height=6cm]{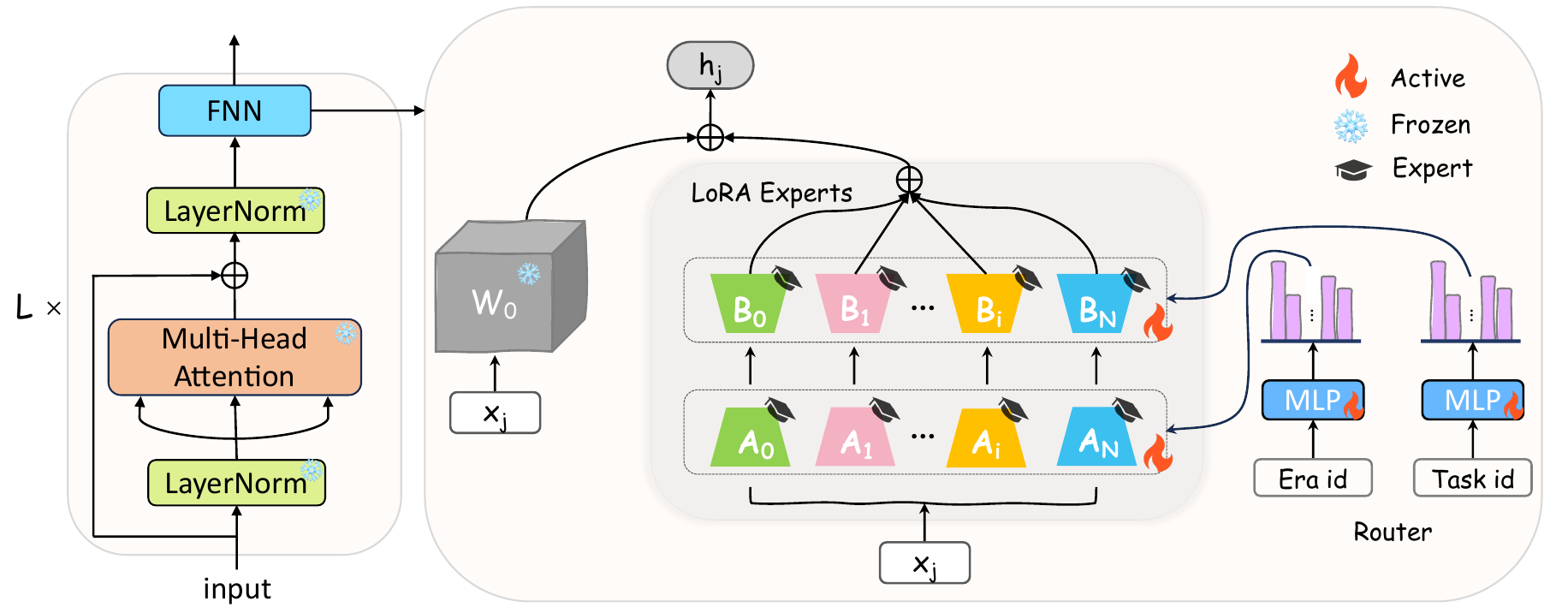}
    \caption{Illustration of Tea-MoELORA framework.}
    \label{fig:enter-label}
\end{figure*}

\subsection{Motivation}

Fine-tuning a single model across heterogeneous tasks and eras often leads to negative interference, which degrades performance.  
Formally, let $\mathcal{T} = \{\mathcal{T}_1, \mathcal{T}_2, \dots, \mathcal{T}_n\}$ denote datasets from different \textit{tasks}, and  
$\mathcal{E} = \{\mathcal{E}_1, \mathcal{E}_2, \dots, \mathcal{E}_m\}$ denote datasets from different \textit{eras}.  
If we optimize one unified model $f_{\theta}$ on the joint distribution

\begin{equation}
\mathcal{L}(\theta) = \sum_{t=1}^{n} \sum_{e=1}^{m} \mathbb{E}_{(x,y) \sim \mathcal{D}_t \times \mathcal{E}_e} \, \ell\!\left(f_{\theta}(x), y\right),
\end{equation}
then the gradient updates from different $(\mathcal{D}_t, \mathcal{E}_e)$ pairs may conflict,  
causing parameter interference and reducing generalization ability.

To mitigate this issue, we draw inspiration from recent advances in multi-task learning, where combining LoRA with MoE has proven highly effective. Specifically, different experts are designated to handle different samples, and the assignment of experts is determined by both the task type and the era of the input.

\subsection{MoELORA}

MoELORA (Mixture-of-Experts with LoRA) is a parameter-efficient fine-tuning framework that combines the advantages of LoRA and the MoE mechanism to better support multi-task and multi-domain scenarios.

LoRA is based on the insight that pre-trained models often lie in a low intrinsic dimension. It decomposes the weight update $\Delta W \in \mathbb{R}^{d_{\text{out}} \times d_{\text{in}}}$ into two trainable low-rank matrices $A \in \mathbb{R}^{r \times d_{\text{out}}}$ and $B \in \mathbb{R}^{d_{\text{in}} \times r}$, where $r \ll d_{\text{in}}, d_{\text{out}}$. The forward process for a linear layer with LoRA is:
\begin{equation}
h = x W_0  + \lambda \cdot x\Delta  W  = x W_0 + \lambda \cdot BAx  
\end{equation}

Here, $W_0$ is the frozen weight from the pre-trained model, $x$ is the input vector, $h$ is the output, and $\lambda$ is a scaling factor.

However, standard LoRA applies a shared parameter update matrix $\Delta W$ across all tasks, which limits its ability to capture task-specific characteristics, particularly in heterogeneous data domains. Our experiments confirm this limitation: when training on multiple datasets jointly, the performance is notably worse than training on each dataset individually. This degradation indicates that, in multi-task and multi-domain scenarios, the shared low-rank subspace can lead to inter-task interference, where conflicting task-specific features hinder each other, resulting in suboptimal performance on certain tasks.

To address this, the MoE mechanism is introduced to enhance model capacity and task adaptability. The parameter efficiency of LoRA with the diversity of MoE experts, aiming to better support the fine-tuning of models across multiple tasks and domains.

~\citet{Liu_Wu_Zhao_Zhu_Xu_Tian_Zheng_2024} introduced MoELORA uses a set of $N$ experts ${\{\mathbb{E}_i\}}_{i=1}^{N}$, where each expert consists of its own pair of low-rank matrices $A_i$ and $B_i$, $\quad A_i \in \mathbb{R}^{\frac{r}{N} \times d_{\text{out}}}$, $B_i \in \mathbb{R}^{d_{\text{in}} \times \frac{r}{N}}$.

During training or inference, a router function computes task-specific weights $\omega_{ji}$ for each expert based on the task $T_j$ that the sample $x_j$ belongs to. The resulting forward process is:
\begin{align}
h_j &= x_j W_0  + \lambda  \cdot \sum_{i=1}^{N} \omega_{ji} \mathbb{E}_i x_j \\
    &= x_j W_0  + \lambda  \cdot \sum_{i=1}^{N} \omega_{ji} B_i A_i x_j
\end{align}

\subsection{Task-era-aware Router}

Historical and modern Chinese exhibit systematic differences in lexical usage, syntactic structures, and discourse organization. For example, classical Chinese often allows verb omission and inverted word order, whereas modern Chinese favors a canonical subject-verb-object order. Relying solely on task-specific routing is therefore insufficient to capture these cross-era linguistic variations. To efficiently share parameters while retaining era-specific knowledge, we propose a sample-level routing mechanism that leverages both task and era metadata, which we refer to as the \emph{Task-Era-Aware Router}.

The router is designed to dynamically combine knowledge from multiple LoRA experts, such that each input is simultaneously guided by task type and temporal domain. Specifically, task-aware gating encourages specialization for semantic objectives, while era-aware gating ensures robustness to diachronic linguistic shifts. By factorizing routing into two orthogonal dimensions, we allow the model to balance cross-task parameter sharing with era-specific adaptation.

First, given an input vector $x_j$, the frozen backbone first produces a base representation:
\begin{equation}
h_0 = x_j W_0^\top,
\end{equation}
where $W_0$ denotes the frozen pretrained weights.  

For each expert $i \in \{1,\dots,N\}$, the LoRA module applies dropout and a low-rank projection:
\begin{equation}
a_i = A_i \, \mathrm{Dropout}(x_j).
\end{equation}
The projection is then modulated by an era-specific weight:
\begin{equation}
\tilde{a}_i = a_i \cdot w_e^i, 
\quad w_e^i \in \mathbb{R}^N,
\end{equation}
which encourages experts to specialize in diachronic linguistic properties.  

Next, the representation is mapped back to hidden space and scaled by a task-specific weight:
\begin{equation}
b_i = B_i \tilde{a}_i, 
\quad \hat{b}_i = w_t^i \cdot \lambda \cdot b_i,
\quad w_t^i \in \mathbb{R}^N,
\end{equation}
where $\lambda$ is the standard LoRA scaling factor.

Finally, we aggregate across all experts:

\begin{equation}
h_j = h_0 + \sum_{i=1}^N \hat{b}_i
\end{equation}

Formally, given an input vector $x_j$, the output of the Task-Era-Aware Gate is defined as:

\begin{equation}
h_j = x_j W_0^\top + \sum_{i=1}^N \Big( w_t^i \cdot \lambda \cdot B_i \big( A_i \, \mathrm{Dropout}(x_j) \cdot w_e^i \big) \Big).
\end{equation}
where $B_i$ and $A_i$ are the parameters of the $i$-th expert, and $w_t^i$ and $w_e^i$ denote the routing weights derived from task and era metadata, respectively. This formulation allows knowledge transfer across all combinations of task- and era-specific experts, enabling greater model expressiveness while maintaining shared low-rank structures.

The routing weights for tasks and eras are computed via linear projections followed by a Softmax:

\begin{equation}
\mathbf{W}_t = \text{Softmax}(W_T v_j^t)
\end{equation}\begin{equation}
\mathbf{W}_e = \text{Softmax}(W_E v_j^e)
\end{equation}

where $v_j^t$ and $v_j^e$ are the feature vectors representing the task and era of the input sample, respectively. Specifically, $v_j^t$ is derived from a learnable task embedding matrix $V_t \in \mathbb{R}^{n \times d_t}$, and $v_j^e$ is derived from a learnable era embedding matrix $V_e \in \mathbb{R}^{m \times d_e}$. Here, $W_T \in \mathbb{R}^{d_t \times N} $ and $W_E \in \mathbb{R}^{d_e \times N} $ denote learnable projection matrices.

We compared two different methods for combining the gates and found that keeping two separate gating mechanisms individually yielded the best performance(as reported in ~\ref{merge_mode}), as it maximally preserves the features from both categories.

\section{Experiments}

\subsection{Experimental Setting}

\textbf{Datasets.} 
We evaluate our model on two Classical Chinese datasets and two Modern Chinese datasets. 
For Classical Chinese, we use the CHED dataset~\cite{Congcong_Zhenbing_Shutan_Wei_Yanqiu_2023} for event classification and the CHisRE dataset~\cite{Tang_Deng_Su_Yang_Wang_2024} for relation extraction. 
For Modern Chinese, we use the ACE 2005 Chinese corpus~\cite{walker2006ace} for event extraction and the DUIE dataset~\cite{Li_He_Shi_Jiang_Liang_Jiang_Zhang_Lyu_Zhu_2019} for relation extraction. 
For DUIE, we randomly sample 10\% of the original training and development sets to construct our training and test sets. 
Statistics of four datasets are summarized in Table~\ref{dataset}.

\textbf{Baselines.} We conduct experiments with three types of baselines.

\textbf{LLM without fine-tuning.} These models are tested using in-context learning (ICL), with two randomly selected samples from each dataset used as demonstrations.

\begin{itemize}
    \item \textbf{Xunzi-Qwen1.5-7B\_chat}~\footnote{https://github.com/Xunzi-LLM-of-Chinese-classics/XunziALLM} is a model built upon Qwen1.5-7B, further trained with classical Chinese texts. It is designed for tasks that involve processing or understanding classical Chinese materials.
    
    \item \textbf{DeepSeek-R1}~\cite{DeepSeek-AI_Guo_Yang_Zhang_Song_Zhang_Xu_Zhu_Ma_Wang_et_al_2025} is an open-source model developed by DeepSeek, supporting a 128K context window, with strong reasoning and generation capabilities.
\end{itemize}

\textbf{LLM with fine-tuning}.
\begin{itemize}
    \item \textbf{LoRA (Single)}: We fine-tune two open-source LLMs, LLaMA-3.1-8B-Instruction, and GLM-4-9b-chat—individually on each dataset using the LoRA strategy.

    \item \textbf{LoRA (Mix)}: We combine the four datasets and fine-tune LLaMA-3.1-8B-Instruction and GLM-4-9b-chat using the LoRA approach.
\end{itemize}  

\textbf{LoRA with MOE Framework.}

\begin{itemize}
    \item \textbf{MoELORA}~\cite{Liu_Wu_Zhao_Zhu_Xu_Tian_Zheng_2024} combines MoE and Low-Rank Adaptation (LoRA) for efficient multi-task fine-tuning. It uses a task-specific gating mechanism to select and combine low-rank expert parameters.
    \item \textbf{TeamLoRA}~\cite{lin-etal-2025-teamlora}. We adopt the core idea of TeamLoRA, originally based on token-level routing, and adapt it to a task-level setting. Specifically, the A matrix is shared across tasks, while the B matrix is split into multiple experts. Experts routing is controlled by both task id and era id, enabling more targeted adaptation.

\end{itemize}

\begin{table}[h]
    \centering
    \setlength{\tabcolsep}{0.8mm}
    \begin{tabular}{c|c|c|c|c}
    \hline
    \Xhline{1.2pt}
         \textbf{Datasets} & \textbf{Training} & \textbf{Dev.} &  \textbf{Test}  & \textbf{Label Num.}\\
    \hline
         CHED& 5,601&1,212&1,216 & 67 \\
    \hline
          CHisRE& 1,560& 194& 194& 12 \\ 
    \hline
        ACE2005 &1,779 & 243 & 246  &33 \\
    \hline
        DUIE & 28,327 &3,402 &3,402 & 48 \\
    \hline
    \Xhline{1.2pt}
    \end{tabular}
    \caption{The details of four IE datasets.}
    \label{dataset}
\end{table}

\begin{table*}[t]
  \centering
  \setlength{\tabcolsep}{0.5mm}
  \begin{tabular}{c|ccc|ccc|cc|cccc}
    \hline
    \Xhline{1.2pt}
    \multirow{3}{*}{\textbf{Model}}
    & \multicolumn{6}{c|}{\textbf{Relation Extraction}} & \multicolumn{6}{c}{\textbf{Event Extraction}}\\
    \cline{2-13}
    & \multicolumn{3}{c}{\textbf{CHisRE}} & \multicolumn{3}{|c|}{\textbf{DUIE}} & \multicolumn{2}{c|}{\textbf{CHED}}  & \multicolumn{4}{c}{\textbf{ACE2005}} \\
    \cline{2-13}
    & P &R&F&P&R&F& Trir-I & Tri-C & Tri-I & Tri-C &Arg-I &Arg-C \\
    \hline
    \multicolumn{13}{c}{LLMs without Fine-tuning}\\
    \hline
    Xunzi-7B &22.73 &30.00 &25.85 &38.55 &54.52&45.16&50.12&18.03&35.73 &25.65 &16.99 &11.81 \\
    \hline
    DeepSeek-R1 &44.91 & 56.00 &49.85 &33.60 &47.83&39.47&67.20&54.89& 50.67&40.54 &33.24&31.63\\
    \hline
    \multicolumn{13}{c}{LLMs with Fine-tuning}\\
      \hline
    Llama-3.1-8B(Single) &57.88 &67.01&62.11 &  74.71&82.59&78.45& \textbf{81.13 }&74.34 &77.66&69.03&63.35&63.90 \\
    \hline
    Llama-3.1-8B(Mix)&51.46&62.81 &56.57&70.60&76.14&73.27&79.28&72.40& 72.20&
    63.73&57.70&59.57\\
    \hline
    GLM-4-9b(Single) & \textbf{65.50}&69.02&67.21& 76.67&70.56&73.49  &81.00 &74.80 &77.61 &67.79 &63.43 &63.30 \\
    \hline

    GLM-4-9b(Mix) & 61.25 & 62.03 &61.64  & \textbf{77.08} & 70.11 & 73.43& 78.49& 73.64 & 75.68&67.78& 57.83 & 57.98\\
    \hline
    \multicolumn{13}{c}{LoRA with MOE}\\
    \hline
    MoELORA &58.80&71.97&64.72& 75.23&\textbf{84.17} &\textbf{79.46} &81.08 &74.24 & 80.55 &71.36 &69.60&70.14 \\
    \hline
    TeamLoRA &59.59 & 71.38 &64.96&75.46&83.42&79.23&80.24&74.42&78.81&69.39&62.61&63.16\\
    \hline
    Ours & 63.67&\textbf{75.81}&\textbf{69.21} & 73.67 & 83.17&78.13 &81.03&\textbf{74.82 }&\textbf{82.43} &\textbf{73.72} &\textbf{70.67} &\textbf{71.77}\\
    
    \Xhline{1.2pt}
  \end{tabular}
  \caption{F1 performance of different models across multiple Chinese information extraction datasets.}
  \label{overall_exp}
\end{table*}
\subsection{Overall Experimental Results}
The experimental results on the four datasets are presented in Table~\ref{overall_exp}, from which we can draw the following observations.

First, the LLMs without fine-tuning perform significantly worse than both the fine-tuned and multi-task learning models. Notably, although Xunzi-7B has undergone continual pretraining on Classical Chinese corpora, it does not demonstrate any clear advantage on the Classical Chinese information extraction datasets. Even DeepSeek-R1 shows unsatisfactory performance across all four information extraction datasets, indicating that information extraction remains a challenging task for LLMs without task-specific fine-tuning.

Second, we compare the performance of LoRA trained separately on each of the four datasets with LoRA trained jointly on the combined datasets. We observe that joint training generally results in inferior performance compared to separate training, particularly for the smaller datasets CHisRE and ACE2005, where the performance drops by approximately 6 points. This highlights the challenge of negative transfer when jointly fine-tuning across heterogeneous datasets.

Finally, compared to LoRA trained jointly without an MoE strategy, methods incorporating MoE demonstrate clear advantages. In fact, all variants of LoRA with MoE outperform jointly trained LoRA. Our proposed Tea-MoELORA further benefits from a more effective router design, leading to substantial improvements across both Classical and Modern Chinese datasets.

\subsection{Ablation Study}
In this section, we validate the effectiveness of the MOE and router components through an ablation study, as shown in Table~\ref{ablation}. First, we observe that incorporating MOE consistently improves performance across the four datasets. Second, when using only the Era information or the Task information as the router signal, the performance is lower compared to using both IDs together.
These results highlight that both MoE and the joint use of task and era signals are essential for achieving the best performance.

\begin{table*}[ht]
  \centering
  \setlength{\tabcolsep}{0.5mm}
  \begin{tabular}{c|ccc|ccc|cc|cccc}
    \hline
    \Xhline{1.2pt}
    \multirow{3}{*}{\textbf{Model}}
    & \multicolumn{6}{c|}{\textbf{Relation Extraction}} & \multicolumn{6}{c}{\textbf{Event Extraction}}\\
  
    \cline{2-13}
    & \multicolumn{3}{c}{\textbf{CHisRE}} & \multicolumn{3}{|c|}{\textbf{DUIE}} & \multicolumn{2}{c|}{\textbf{CHED}}  & \multicolumn{4}{c}{\textbf{ACE2005}} \\
    \cline{2-13}
    & P &R&F&P&R&F& Trir-I & Tri-C & Tri-I & Tri-C &Arg-I &Arg-C \\
     \hline
     W/o MOE  &61.25 & 62.03 &61.64  & 77.08 & 70.11 & 73.43& 78.49& 73.64 & 75.68&67.78& 57.83 & 57.98\\
     \hline
    W/o Era id &58.80&71.97&64.72& 75.23&84.17 &79.46 &81.08 &74.24 & 80.55 &71.36 &69.60&70.14\\
    \hline
    W/o Task id &61.60&72.03 &66.40&75.30&84.89&\textbf{79.81}&\textbf{81.45} &\textbf{75.45}&\textbf{82.84} &\textbf{73.88}&65.84 &.66.67\\
    \hline
    Ours & 63.67&75.81&\textbf{69.21} & 73.67 & 83.17&78.13 &81.03&74.82 &82.43 &73.72 &\textbf{70.67} &\textbf{71.77}
    \\
    \hline
    \Xhline{1.2pt}
    \end{tabular}
  \caption{Ablation study.}
  \label{ablation}
\end{table*}
\begin{table*}
  \centering
  \setlength{\tabcolsep}{0.5mm}
  \begin{tabular}{c|ccc|ccc|cc|cccc}
    \hline
    \Xhline{1.2pt}
    \multirow{3}{*}{\textbf{Model}}
    & \multicolumn{6}{c|}{\textbf{Relation Extraction}} & \multicolumn{6}{c}{\textbf{Event Extraction}}\\
  
    \cline{2-13}
    & \multicolumn{3}{c}{\textbf{CHisRE}} & \multicolumn{3}{|c|}{\textbf{DUIE}} & \multicolumn{2}{c|}{\textbf{CHED}}  & \multicolumn{4}{c}{\textbf{ACE2005}} \\
    \cline{2-13}
    & P &R&F&P&R&F& Trir-I & Tri-C & Tri-I & Tri-C &Arg-I &Arg-C \\
    \hline
    Two task id& 60.89 &70.01 &65.13 &72.93&83.86 &78.01&80.80&74.74&83.33 &73.27 &67.26 &68.75\\
    \hline
     Four task id& 63.67&75.81&69.21 & 73.67 & 83.17&78.13 &81.03&74.82 &82.43 &73.72 &70.67 &71.77
     \\
    \hline
    \Xhline{1.2pt}
    \end{tabular}
  \caption{Results under different levels of task granularity.}
  \label{task_id_grain}
\end{table*}


\subsection{Effect of Task Granularity}

In this section, we investigate the effect of task ID granularity. Specifically, we conduct experiments on two RE datasets and two EE datasets. In the MOE router, we compare assigning a single ID to all RE tasks and another ID to all EE tasks (“Two task IDs”) with assigning a distinct ID to each dataset (“Four task IDs”). The results are presented in Table~\ref{task_id_grain}. According to the experimental results, even tasks of the same type exhibit significant differences when their labels and schemas differ. This also leads to variations in how experts are utilized, further supporting the need to treat each dataset as an individual task.

\subsection{Effect of Router Signal Merge Strategy}
\label{merge_mode}

\begin{figure}[h]
    \centering
    \small
    \includegraphics[width=1\linewidth]{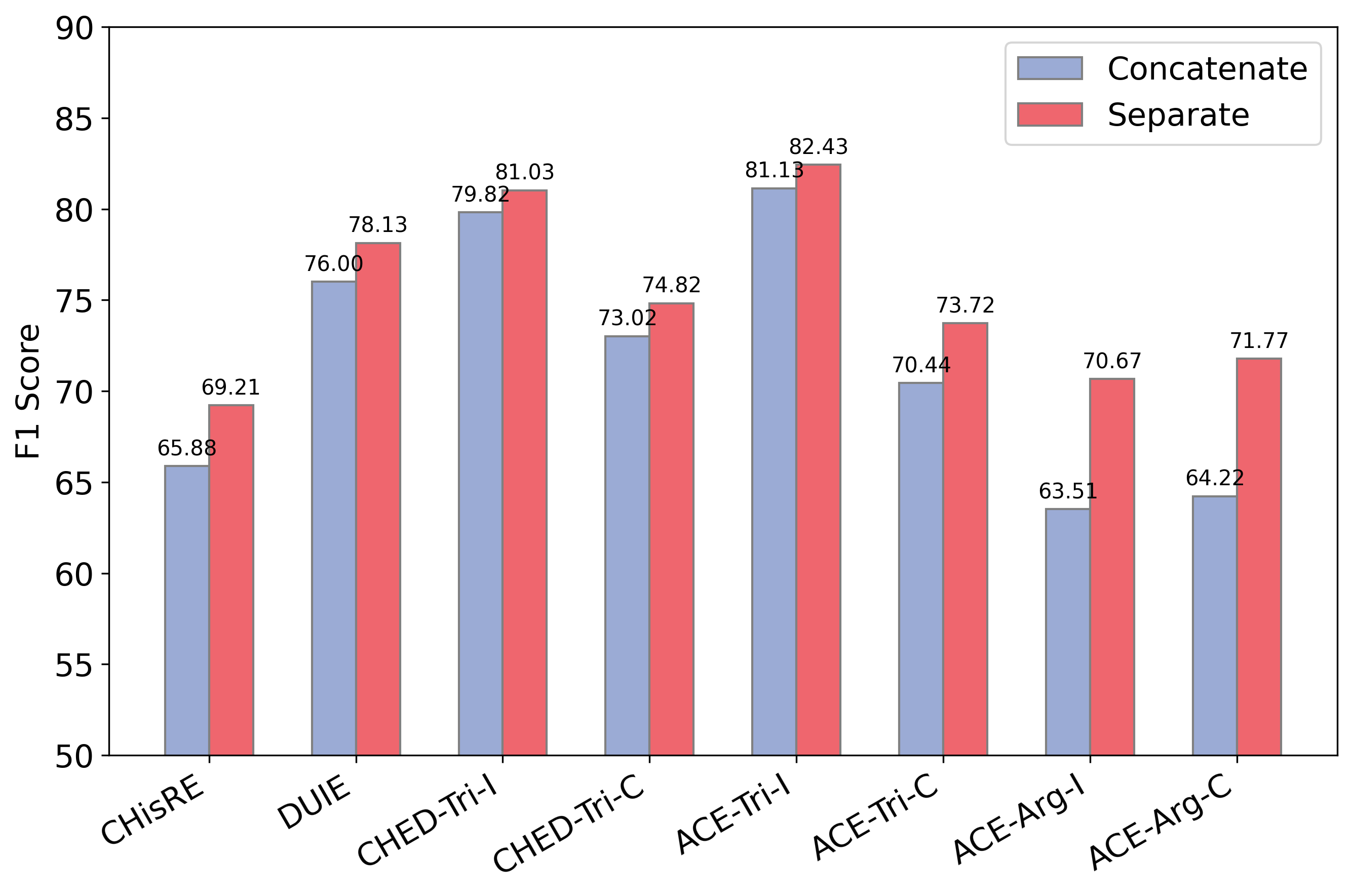}
    \caption{Results with different strategies for integrating era and task signals.}
    \label{gate_compare}
\end{figure}

In this section, we evaluate different strategies for integrating era and task signals. The results are illustrated in the bar chart shown in Figure~\ref{gate_compare}. Compared to simply concatenating both signals and feeding them into a single MLP-based router, using separate routers—i.e., assigning one router to each signal type—achieves better performance. This demonstrates that using multiple routers is particularly effective for datasets with rich characteristics, where differences arise not only from task type but also from temporal or stylistic variations.

\subsection{Expert Utilization Preferences}

\begin{figure}
    \centering
    \centering
    \subfigure[Era weight distribution form Team-MoELoRA] {
     \label{f3a}     
    \includegraphics[width=1\columnwidth]{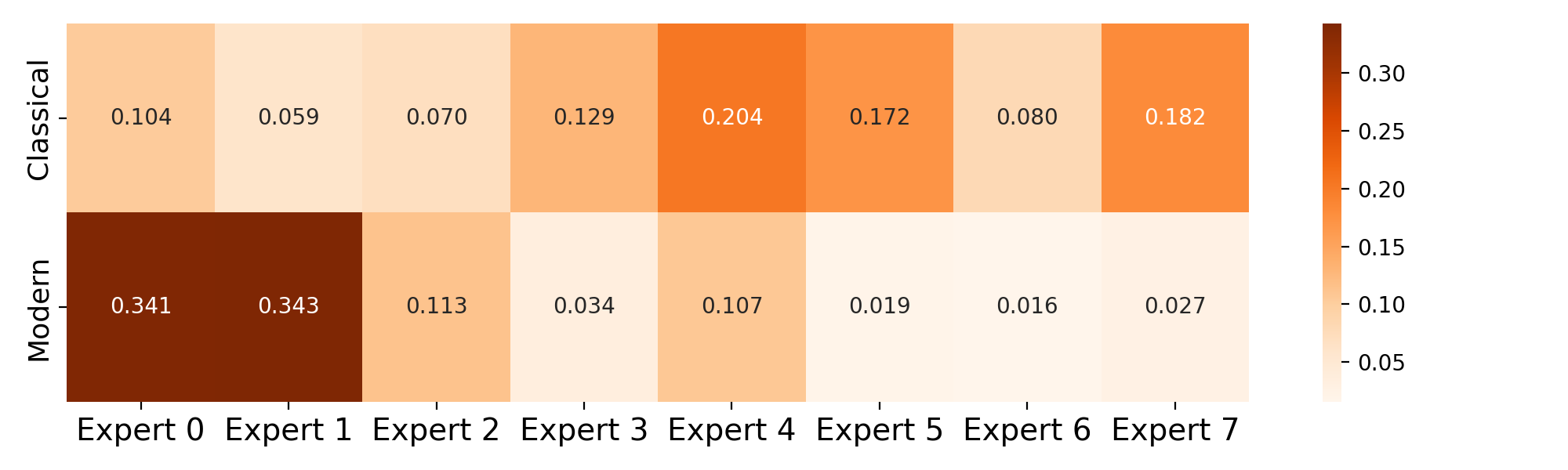}  
    } 
    \centering
    \subfigure[Era weight distribution (using only Era ID)] {
     \label{f3b}     
    \includegraphics[width=1\columnwidth]{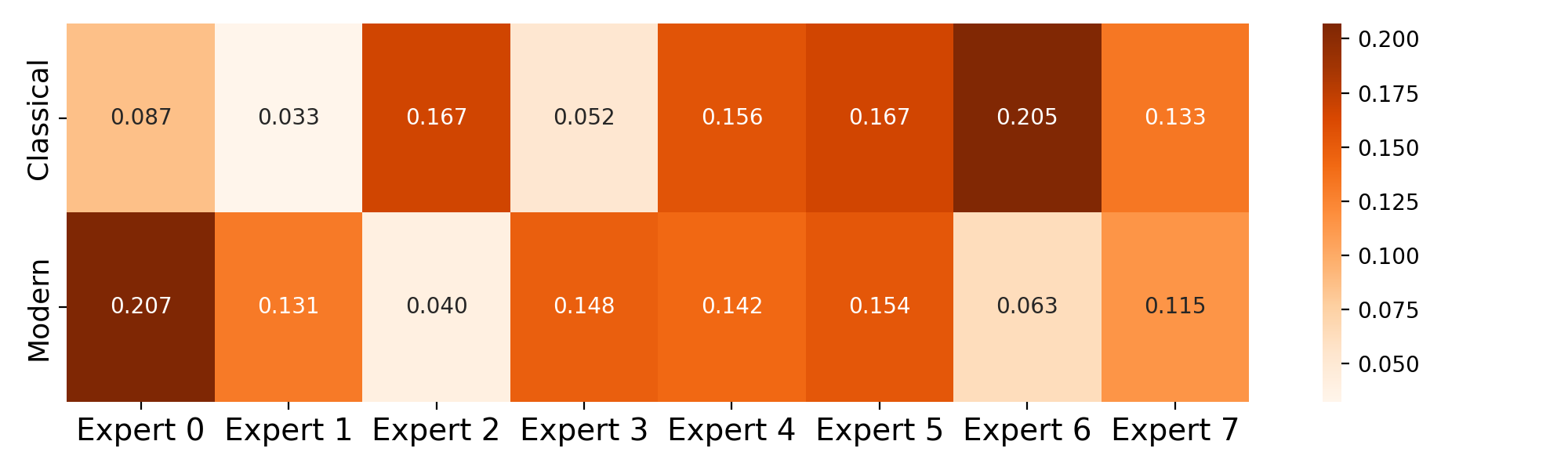}  
    } 
        \centering
    \subfigure[Task weight distribution form Team-MoELoRA] {
     \label{f3c}     
    \includegraphics[width=1\columnwidth]{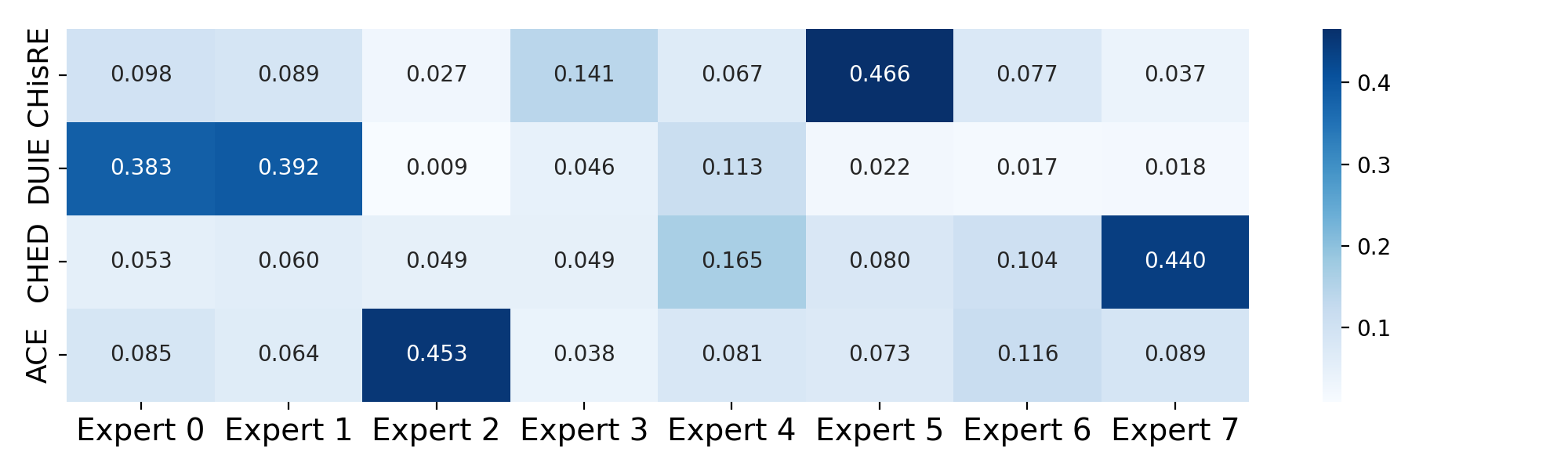}  
    } 
    \centering
    \subfigure[Task weight distribution (using only Task ID)] {
     \label{f3d}     
    \includegraphics[width=1\columnwidth]{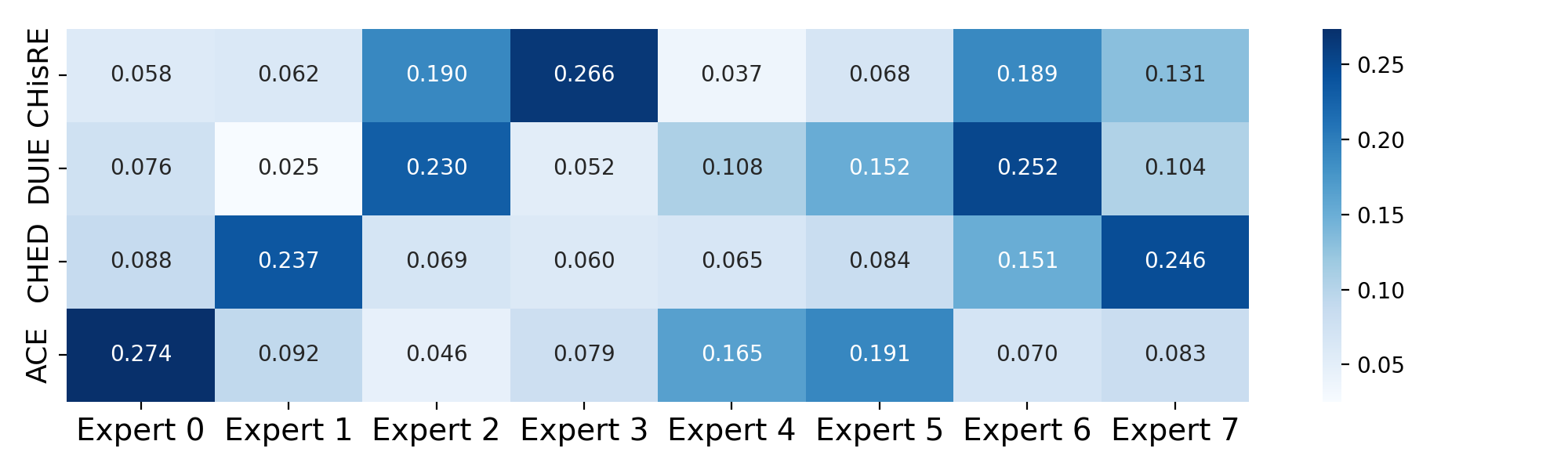}  
    }
    \caption{Visualization of expert weight distributions for two types of information.}
    \label{weight_districution}
\end{figure}

\begin{figure*}[t]
    \centering
    \includegraphics[width=1\linewidth]{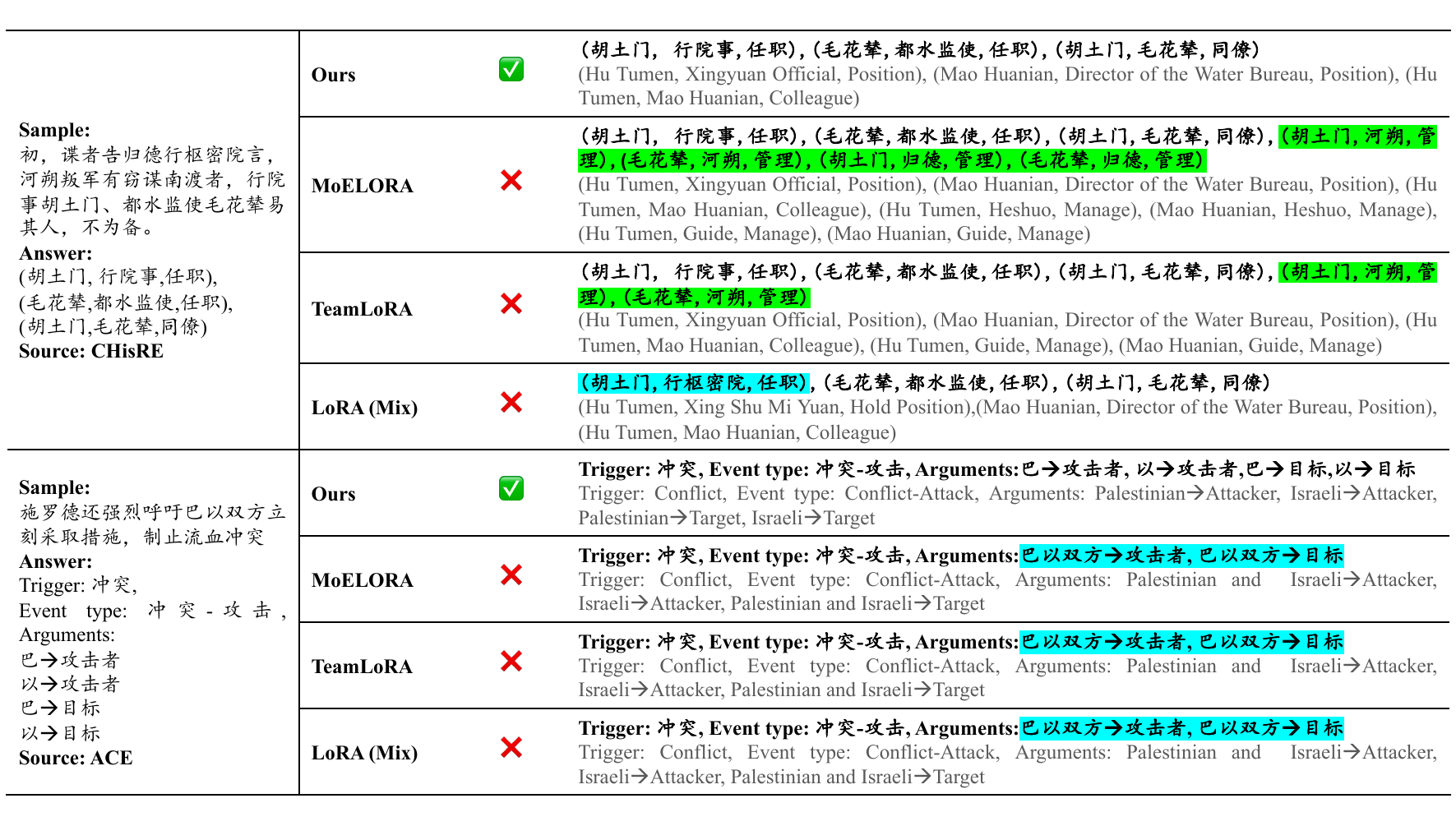}
    \caption{Case study of relation extraction and event detection. Green highlights indicate hallucinated predictions, while blue highlights denote incorrect predictions.}
    \label{case}
\end{figure*}

We present the visualization of expert weight distribution in Figure~\ref{weight_districution}. 
According to Figure~\ref{f3a}, the results demonstrate that samples from different eras are indeed assigned to different experts. Specifically, Classical Chinese samples are predominantly assigned to Expert 4 and Expert 7, whereas Modern Chinese samples are mainly assigned to Expert 0 and Expert 1. For each task, Figure~\ref{f3d} shows that different tasks emphasize different experts. Specifically, CHisRE mainly relies on Expert 5, DUIE primarily attends to Expert 0 and Expert 1, CHED is focused on Expert 7, and ACE is assigned mainly to Expert 2. These results indicate that both era and task factors guide the model to make distinct expert selections, confirming the effectiveness of the expert specialization mechanism.

Moreover, by comparing Figures~\ref{f3a}--\ref{f3b}, ~\ref{f3c}--\ref{f3d} and the smoothness metrics (variance, entropy, and max-min difference as shown in Appendix~\ref{smooth}), we observe that the distribution for models using only Era ID or Task ID is generally smoother and more balanced across experts, while the combined Era and Task ID model shows more concentrated assignments, indicating stronger specialization when both factors are considered.


\subsection{Case Study}

We present two representative examples to illustrate the performance differences among Tea-MoELORA(Ours), MoELORA, TeamLoRA, and LoRA (Mix) on relation extraction and event detection.

In the first example, from CHisRE, the sentence mentions the officials ``胡土门''(Hutumen) and ``毛花辇'' (Maohuanian), and their respective positions. Our model correctly predicts all three relations: the official positions of ``胡土门'' and ``毛花辇'', and their colleague relationship. Other models either hallucinate additional relations, such as management relations involving places ``河朔 (Heshuo)''or incorrectly link ``胡土门 (Hutumen)'' to other entities, demonstrating issues with over-prediction or missing entity constraints.

In the second example, from ACE, the sentence describes a conflict between Palestinian and Israeli parties. Our model accurately extracts the trigger ``冲突'' (Conflict), identifies the event type as ``冲突-攻击'' (Conflict-Attack), and assigns correct roles to the participants. In contrast, MoELORA, TeamLoRA, and LoRA (Mix) incorrectly merge the participants as a single entity (``巴以双方 (both Palestinian and Israeli)''), resulting in inaccurate argument assignments.

These examples highlight that our model not only reduces hallucinated relations but also better preserves entity-level accuracy and role assignments, especially in complex historical or conflict-related sentences.

\section{Conclusion}
In this study, we propose the Tea-MoELORA framework for information extraction across classical and modern Chinese texts, which effectively combines the strengths of LoRA and MOE. Considering both task-specific characteristics and era-related features of the corpus, the MOE router incorporates two types of information: task and era. Experimental results demonstrate that our approach is more effective than directly training LoRA on the mixed corpus.

\bibliography{custom}

\clearpage
\appendix
\section{Appendix A}
\label{smooth}

\begin{table}[h!]
\centering
\small
\setlength{\tabcolsep}{0.05mm}
\begin{tabular}{lccc}
\hline
\textbf{Model} & \textbf{Variance} & \textbf{Entropy} & \textbf{Max-Min Diff} \\
\hline
Task ID (Tea-MoELoRA) & 0.01825 & 1.62666 & 0.40703 \\
Era ID (Tea-MoELoRA) & 0.00981 & 1.78265 & 0.23610 \\
Task ID (only) & 0.00557 & 1.90709 & 0.21752 \\
Era ID (only) & 0.00286 & 1.97587 & 0.16970 \\
\hline
\end{tabular}
\caption{Comparison of Heatmap Smoothness Across Different Models. Higher entropy and lower variance/Max-Min indicate a smoother distribution.}
\label{tab:heatmap_smoothness}
\end{table}

\end{CJK}
\end{document}